# Contaminant source identification in groundwater by means of Artificial Neural Network

Daniele Secci*, Laura Molino, Andrea Zanini

Department of Engineering and Architecture, University of Parma, Parco Area delle Scienze 181/A, 43124 Parma, Italy

* Corresponding author, daniele.secci@unipr.it

**Abstract**

In a desired environmental protection system, groundwater may not be excluded. In addition to the problem of over-exploitation, in total disagreement with the concept of sustainable development, another not negligible issue concerns the groundwater contamination. Mainly, this aspect is due to intensive agricultural activities or industrialized areas. In literature, several papers have dealt with transport problem, especially for inverse problems in which the release history or the source location are identified. The innovative aim of the paper is to develop a data-driven model that is able to analyze multiple scenarios, even strongly non-linear, in order to solve forward and inverse transport problems, preserving the reliability of the results and reducing the uncertainty. Furthermore, this tool has the characteristic of providing extremely fast responses, essential to identify remediation strategies immediately. The advantages produced by the model were compared with literature studies. In this regard, a feedforward artificial neural network (ANN), which has been trained to handle different cases, represents the data-driven model. Firstly, to identify the concentration of the pollutant at specific observation points in the study area (forward problem); secondly, to deal with inverse problems identifying the release history at known source location (also in the case with multiple sources); then, identifying the release history and, at the same time, the location of one source in a specific sub-domain of the investigated area. At last, the observation error is investigated and estimated. The results are satisfactorily achieved, highlighting the capability of the ANN to deal with multiple scenarios by approximating nonlinear functions without the physical point of view that describes the phenomenon, providing reliable results, with very low computational burden and uncertainty.

# 1 Introduction

Groundwater contamination is an environmental problem that is increasing rapidly and need attention. Unfortunately, often it is necessary to treat the water in order to be suitable for drinking. A sustainable management could protect the quality of groundwater rather than developing expensive treatment systems (Katsanou and Karapanagioti, 2017). In order to prevent the spread of a pollutant in aquifer, knowledge of the location of the contaminant source and its release over time becomes of fundamental importance. Indeed, identifying the source of contaminants is a problem that has attracted great attention over the past four decades, as highlighted in recent reviews by Gómez-Hernández and Xu (2021) and Barati Moghaddam et al. (2021).

Contaminant source identification starting from few concentration observations is a good representative of an inverse problem in hydrology. In literature, to solve this kind of problem, a variety of approaches exist, and it is possible to classify the inverse methods to numerically solve the problem of contaminant source identification in three main areas (Barati Moghaddam et al., 2021): mathematics-based, stochastic-based, and optimization-based. Mathematics-based methods directly tackle to an inverse source problem using numerical or analytical techniques. Two common ways to handle stability issue are regularization and stabilization methods (Skaggs and Kabala, 1994; Liu and Ball, 1999). The stochastic approaches (Woodbury and Ulrich, 1998; Butera et al., 2013; Cupola et al. 2015; Gzyl et al. 2015; Zanini and Woodbury, 2016; Xu and Gómez-Hernández, 2016; Todaro et al., 2021; Wang et al., 2021; Wang et al., 2022) consider the problem in a stochastic framework and the parameters to estimate become random variables. The optimization-based approaches consist of the integration of both simulation and optimization models. The simulation model solves the flow and transport equations for given initial and boundary conditions. Then, the differences between simulated and observed data are minimized through an optimization algorithm (Ayvaz, 2010; Jamshidi et al. 2020). The reader is referred to Gómez-Hernández and Xu (2021) and to Barati

Moghaddam et al. (2021) for extensive reviews of the source reconstruction problem in groundwater hydrology and groundwater-surface hydrology, respectively.

Over the past decade, Artificial Neural Networks (ANNs) have grown very rapidly, thanks to improved computing power and technology. ANNs were firstly applied to contaminant source reconstruction by Singh and Datta (2004) and Singh et al. (2004). They considered the concentrations observed at monitoring points as input to the network and the release history at the contaminant source as the output. Singh and Datta (2004) developed an ANN-based methodology to simultaneously solve the problems of groundwater pollution source identification and hydro-dispersive parameters of the aquifer. Singh et al. (2004) investigated the efficacy of ANN considering multiple source and noise on observations. Chaubey and Srivastava (2020) proposed an application of ANN to estimate the source location and release concentration using a 1D simple study case. Ayaz (2021) proposed an ANN to estimate the release history of groundwater pollution source without information about the starting time of the release. Recently, Pan et al. (2022) proposed a deep residual neural network as a forward surrogate model combined with an ensemble smoother particle filter in order to estimate the groundwater contamination source together with the aquifer hydraulic conductivity.

This work proposes an application of ANN to contaminant source reconstruction with the objective of minimizing the training period and the information required in the inverse procedure. In order to reduce the number of the training set the Latin hypercube sampling (McKay et al., 1979) was considered. Furthermore, to reach the solution in an acceptable way for the applications carried out, it is necessary to calibrate the network and its parameters adequately. The black box structure of ANNs allows to consider multiple scenarios by approximating any type of function, also strongly non-linear, without the physical point of view that describes the phenomenon. This leads to consider new scenarios never analyzed before in this field and, at the same time, to reduce the computational cost together with the number of observations necessary to implement the model. In the present work ANNs have been used to estimate the location of the source and its release over time, starting from few concentrations observed at monitoring wells. To evaluate and compare the proposed procedure, a complex literature case study was considered. The procedure is very efficient and allows to considerably reduce the number of observations to solve the inverse problem. Furthermore, for the first time, ANNs were used to simultaneously estimate release history and error on observations. The literature test case was proposed by Ayvaz (2010) and later adopted by Xing et al. (2019) and Jamshidi et al. (2020). The test case consists of a heterogeneous aquifer with multiple contaminant sources and seven monitoring wells.

The test case considered is investigated considering different scenarios and objectives:

1. estimation of the pollutant concentration in monitoring wells with a release source (direct problem);
2. estimation of the release history at one contaminant source with known location;
3. estimation of the release history at two contaminant sources with known location;
4. simultaneous estimation of the release history and location of contaminant source with unknown location;
5. simultaneous estimation of the release history of two sources with known location and of the error on observations.

Objectives 2, 3 and 4 were investigated corrupting the observations with different errors.

The paper is organized as follows. In Section 2, the methodologies adopted to implement the data-driven model and its application for solving a literature case study dealing with several forward and inverse scenarios are presented. Then, Section 3 shows the main results that are discussed in Section 4 together with the Conclusions.

## 2  Material and Methods

### 2.1  Groundwater Flow and Transport

To study contaminant transport problems in the aquifer, it is necessary to know the flow field, in particular, the velocity field. In the following, for the sake of simplicity, a confined aquifer, with known hydraulic parameters, characterized by a two-dimensional flow equation (Eq. 1) is considered; however, the adopted procedure can be applicable also for unconfined and three-dimensional aquifers. Including the Darcy law, Eq.1 shows (in cartesian coordinate $\boldsymbol{\xi} = (\zeta, \eta)$ assumed to coincide with the principal directions of the symmetric tensor representative of the transmissivity) the mass balance with regard to a heterogeneous and anisotropic confined aquifer:

$$\frac{\partial}{\partial \zeta}\left(T_{\zeta\zeta}(\zeta,\eta)\frac{\partial h}{\partial \zeta}(\zeta,\eta,t)\right) + \frac{\partial}{\partial \eta}\left(T_{\eta\eta}(\zeta,\eta)\frac{\partial h}{\partial \eta}(\zeta,\eta,t)\right) = S\frac{\partial h}{\partial t}(\zeta,\eta,t) + Q(\zeta,\eta,t) \qquad (1)$$

where $T_{\zeta\zeta}$ and $T_{\eta\eta}$ $[L^2T^{-1}]$ represent the principal values of transmissivity along the directions $\zeta$ and $\eta$, $t$ $[T]$ is the time, $h$ is the piezometric head $[L]$, $Q$ $[LT^{-1}]$ is the flow rate entered or extracted per unit area sources (positive if entering) and $S$ $[-]$ is the storativity of the porous medium, that represents the volume of water released per unit of planimetric area of the aquifer due to a unit lowering of the piezometric head. Assuming that the contaminant performs as a tracer, it is possible to solve the contaminant problem in two steps: first to solve the flow problem, then to solve the transport problem on the know flow field having assumed specific boundary and initial conditions. Eq. 2 defines the transport process with regard to an injection of non-reactive and non-sorbing solute at a point source,

$$\frac{\partial(\phi C(\boldsymbol{\xi},t))}{\partial t} = \nabla \cdot [\phi \mathbf{D}(\boldsymbol{\xi})\nabla C(\boldsymbol{\xi},t)] - \nabla \cdot [\phi \mathbf{u}(\boldsymbol{\xi},t)C(\boldsymbol{\xi},t)] + s(\boldsymbol{\xi}_0,t)\delta(\boldsymbol{\xi} - \boldsymbol{\xi}_0) \qquad (2)$$

where $\boldsymbol{\xi}$ is the position vector of the point location in the two-dimensional aquifer, $\boldsymbol{\xi}_0$ is the location of the source, $C(\boldsymbol{\xi},t)$ $[ML^{-3}]$ is the concentration at specific location $\boldsymbol{\xi}$ and time $t$ $[T]$, $\phi$ $[-]$ is the effective porosity, $\mathbf{u}(\boldsymbol{\xi},t)$ $[LT^{-1}]$ is the effective velocity vector field at specific location $\boldsymbol{\xi}$ and time $t$ $[T]$, $\boldsymbol{D}(\boldsymbol{\xi})$ $[L^2T^{-1}]$ is the dispersion tensor, $\nabla$ is the differential operator Nabla in the spatial coordinates $\boldsymbol{\xi}$, $s(\boldsymbol{\xi}_0,t)$ $[MT^{-1}]$ is the released mass rate of contaminant per unit time injected into the aquifer through the source and $\delta$ $[L^{-3}]$ is the Dirac delta function.

## 2.2 Artificial Neural Network

Neural networks represent information processing systems inspired by the functioning of biological nervous systems. These are formed by simple units (neurons) interconnected with each other through 'synaptic connections', which are considered as weights within the neural networks. Some of these units receive inputs from the external environment, others produce outputs that return to the environment and others, if any, exchange information within the network itself. In this sense, three layers define ANNs: input layer, output layer and hidden layers (Hagan et al., 1996).

Since the neural networks could be composed of more than a single neuron and equipped with several synaptic connections, it is appropriate to process the system through a vector notation in which the components are real numbers; the generic signal ($\boldsymbol{\omega}$) emitted by a layer (hidden or output) which encodes information through numbers, turns out to be:

$$\boldsymbol{\omega} = g(\boldsymbol{Wa} + \boldsymbol{b}) \qquad (3)$$

where $\boldsymbol{a}$ is an input vector for that layer, $\boldsymbol{W}$ is a weight matrix in which the entry $w_{i,j}$ in the $i$-th row and in $j$-th column is related to the receiving neuron $i$ and to the $j$-th component $a_j$ of the input vector $\boldsymbol{a}$, and $\boldsymbol{b}$ is a bias term vector in which the $i$-th component is related to the receiving neuron $i$. The activation function $g$ processes the input information and can take various forms. The most used activation functions are the sigmoid or hyperbolic tangent for the input-hidden layers and the identity or piecewise nonnegative identity function (ReLu, Rectified Linear Unit) for the hidden-output layers. Since the term $g(\boldsymbol{Wa} + \boldsymbol{b})$ represents the output of the downstream layer, the values of the net weights and biases determine what will be the response of the whole system. With the aim to develop a data-driven model that starts from the initial information (input) and goes to the desired response (output), the computational process modifies the weights and the biases through specific learning algorithms in order to obtain the desired output. As in any optimization process, a Loss Function ($L$) is defined, which represents the measure of the error that the neural network, interpreted as a differentiable system, makes on a specific dataset and with respect to all the matrices of the weights $\boldsymbol{W}$ and to all the bias terms $\boldsymbol{b}$ which becomes parameters for that differentiable system; in short, it is possible to describe the change of the error as a function of the change of the weights and the biases. The $L$ function depends only on the weights and biases of the net; therefore, the gradient of that function (which represents the maximum growth direction of the function) is defined by a vector of partial derivatives of L with regard to each weight and bias. According to the fact that the Loss Function wants to be reduced and not increased, it is necessary to modify the weights and biases in the opposite direction to the gradient of L. The algorithm used for multi-layer networks (with at least one hidden layer) to compute the derivatives is the back-propagation algorithm in which the derivatives are computed from the last layer to the first. In the following paragraphs, the network architecture, the characteristic parameters and the used learning algorithm to update the parameters will be described in detail.

## 2.3 Architecture of the network

The network is composed by three layers: input, hidden and output. Each layer is a Euclidean vector space. Considering $d_1, d_2, d_3$ the dimensions of such vector spaces, it means that input data are real vectors with $d_1$ components, output data are real vectors with $d_3$ components and in the hidden layer $d_2$ neurons are considered.

From a functional analysis point of view the network, interpreted as a differentiable system, is simply a composition of multivariable vector-valued functions: affine transformations $f$ and linear or nonlinear functions $g$ (activation functions), from $\mathbb{R}^{d_1}$ in $\mathbb{R}^{d_3}$.

$$\mathbb{R}^{d_1} \xrightarrow{f_1} \mathbb{R}^{d_2} \xrightarrow{g_1} \mathbb{R}^{d_2} \xrightarrow{f_2} \mathbb{R}^{d_3} \xrightarrow{g_2} \mathbb{R}^{d_3} \tag{4}$$

In this framework any finite dataset with $N$ data can be considered as an indexed family $\{(\boldsymbol{x}^{(i)}, \boldsymbol{y}^{(i)})\}_{i=1,\dots,N} \subset \mathbb{R}^{d_1} \times \mathbb{R}^{d_3}$ of ordered couples where, each $\boldsymbol{y}^{(i)}$ is the vector target corresponding to the arbitrary input data $\boldsymbol{x}^{(i)}$. With the aim to analyse the analytic expressions of the functions of the network individually, for future convenience superscripts to better identify layers and their dimensions will be used, and the independent variable $\boldsymbol{x}$ of the first space $\mathbb{R}^{d_1}$ will be denoted as $\boldsymbol{x} = \boldsymbol{a}^{(1)}$. The first function of the network is an affine transformation:

$$f_1(\boldsymbol{a}^{(1)}) = \boldsymbol{W}^{(1)} \boldsymbol{a}^{(1)} + \boldsymbol{b}^{(1)} \tag{5}$$

where $\boldsymbol{W}^{(1)} \in M_{d_2 \times d_1}(\mathbb{R})$ is a matrix of real numbers (weights) with $d_2$ rows and $d_1$ columns, $\boldsymbol{b}^{(1)} \in M_{d_2 \times 1}(\mathbb{R})$ is a column vector (biases) with $d_2$ rows and 1 column, $\boldsymbol{a}^{(1)} \in M_{d_1 \times 1}(\mathbb{R})$ is a column vector with $d_1$ rows and 1 column and the operations of multiplication and addition are the usual ones in the algebra of matrices. Let's denote by $\mathbf{z}^{(2)} = f_1(\mathbf{a}^{(1)})$ the image of the first function, where $\mathbf{z}^{(2)} \in \mathbb{R}^{d_2}$. The second function of the composition is $g_1$, the first activation function of the network, and it could be defined for example using the hyperbolic tangent function on each component as:

$$g_1\left(z_1^{(2)}, \ldots, z_{d_2}^{(2)}\right) = \left(\frac{2}{1+\exp\left(-2z_1^{(2)}\right)} - 1, \ldots, \frac{2}{1+\exp\left(-2z_{d_2}^{(2)}\right)} - 1\right) \tag{6}$$

Note that such $g_1$ shrinks every value of $\mathbf{z}^{(2)}$ into the open interval $(-1,1)^{d_2}$. The input data for the output layer has been denoted as $\mathbf{a}^{(2)} = g_1(\mathbf{z}^{(2)})$. The third function of the network is still an affine transformation:

$$f_2(\mathbf{a}^{(2)}) = \mathbf{W}^{(2)}\mathbf{a}^{(2)} + \mathbf{b}^{(2)} \tag{7}$$

where $\mathbf{W}^{(2)} \in M_{d_3 \times d_2}(\mathbb{R})$ is a matrix of real numbers (weights) with $d_3$ rows and $d_2$ columns, $\mathbf{b}^{(2)} \in M_{d_3 \times 1}(\mathbb{R})$ is a column vector (biases) with $d_3$ rows and 1 column, $\mathbf{a}^{(2)} \in M_{d_2 \times 1}(\mathbb{R})$ is a column vector with $d_2$ rows and 1 column; $\mathbf{z}^{(3)} = f_2(\mathbf{a}^{(2)})$ is the image of the third function, where $\mathbf{z}^{(3)} \in \mathbb{R}^{d_3}$. The last function of the composition is $g_2$, the second activation function of the network and it could be defined, for example, as:

$$g_2\left(z_1^{(3)}, \ldots, z_{d_3}^{(3)}\right) = \left(z_1^{(3)}, \ldots, z_{d_3}^{(3)}\right) \tag{8}$$

In this case $g_2$ is the identity function of $\mathbb{R}^{d_3}$. The network, written as a composition of functions, gives the following output:

$$\mathbf{h}_{W,b}(\mathbf{x}) = g_2\left(\mathbf{W}^{(2)} g_1\left(\mathbf{W}^{(1)}\mathbf{x} + \mathbf{b}^{(1)}\right) + \mathbf{b}^{(2)}\right) \tag{9}$$

which depends on the independent variable $\mathbf{x}$, the matrices $\mathbf{W}^{(1)}$, $\mathbf{W}^{(2)}$ and the biases terms $\mathbf{b}^{(1)}$, $\mathbf{b}^{(2)}$ parameters for the resulting composed function. Processing the function $\mathbf{h}_{W,b}$ on a dataset $\{(\mathbf{x}^{(i)}, \mathbf{y}^{(i)})\}_{i=1,\ldots,N}$ is possible to evaluate the error between $\mathbf{h}_{W,b}(\mathbf{x}^{(i)})$ and the related target $\mathbf{y}^{(i)}$ as $\|\mathbf{h}_{W,b}(\mathbf{x}^{(i)}) - \mathbf{y}^{(i)}\|$, where $\|\cdot\|$ is the Euclidean norm in $\mathbb{R}^{d_3}$. In Fig. 1, the architecture of the neural network is reported.

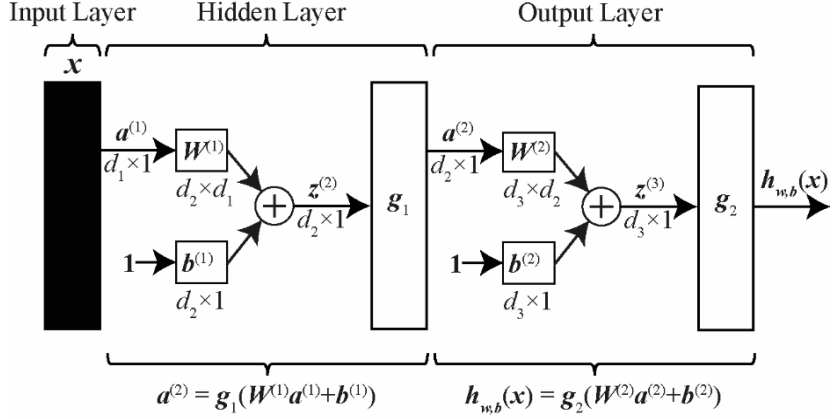

Fig. 1. Sketch view of the network

For future convenience the resulting composed function $h_{W,b}$, which globally describe the network, will be denoted as $h_\Theta$ in which $\Theta = (\theta_1, \ldots, \theta_n) \in \mathbb{R}^n$ is an ordered vector of $n$ components and where $n = d_1 d_2 + d_2 d_3 + d_2 + d_3$ is the numbers of elements in $W^{(1)}, W^{(2)}, b^{(1)}, b^{(2)}$.

## 2.4 Loss Function-MSE

The neural network performance index is a function of all the network parameters $\Theta = (\theta_1, \ldots, \theta_n)$ (weights and biases) and it is defined:

$$L(\Theta) = \frac{1}{Nd_3}\sum_{i=1}^{N}\sum_{j=1}^{d_3}\left(h_\Theta(x^{(i)})_j - y_j^{(i)}\right)^2 = \frac{1}{Nd_3}\sum_{i=1}^{N}\left\|h_\Theta(x^{(i)}) - y^{(i)}\right\|^2 \tag{10}$$

- $N$ is the number of ordered couples in the dataset
- $(x^{(i)}, y^{(i)}) \in \mathbb{R}^{d_1} \times \mathbb{R}^{d_3}$ is the $i$-th couple ordered of the dataset where $y^{(i)}$ is the vector target corresponding to the input data $x^{(i)}$ and $i \in \{1, \ldots, N\}$
- $h_\Theta(x^{(i)})$ is the response of the network computed on the $i$-th input data $x^{(i)}$ and it depends on all the network parameters $\Theta$
- $\|-\|$ is the Euclidean norm in $\mathbb{R}^{d_3}$
- $\|h_\Theta(x^{(i)}) - y^{(i)}\|$ is the distance, in the Euclidean metric space $\mathbb{R}^{d_3}$, between the response of the network and its target computed on the $i$-th input data $x^{(i)}$ and it represents the magnitude of the $i$-th error $e_i(\Theta)$.

The aim of network training is to optimize the loss function $L$ by searching for those parameters $\Theta$ that minimize $L$ making the error small on the whole given dataset. The algorithm is iterative.

## 2.5 Levenberg-Marquardt algorithm

The Levenberg-Marquardt numerical optimization technique, which can be considered as a modification of Newton's method, is very well suited in those neural networks in which the performance index is defined by the mean squared error of nonlinear functions. Note that the Levenberg-Marquardt algorithm, as well as the Gauss-Newton method, does not require calculation of second derivatives, reducing the computational cost during the training phase.

For the sake of brevity, the reader is directed to (Hagan and Menhaj, 1994) for details that lead to the Levenberg-Marquardt algorithm:

$$\boldsymbol{\Theta}^{(k+1)} = \boldsymbol{\Theta}^{(k)} - \left(\boldsymbol{J}^T\big(\boldsymbol{\Theta}^{(k)}\big)\boldsymbol{J}\big(\boldsymbol{\Theta}^{(k)}\big) + \mu_k \boldsymbol{I}\right)^{-1}\boldsymbol{J}^T\big(\boldsymbol{\Theta}^{(k)}\big)\boldsymbol{e}\big(\boldsymbol{\Theta}^{(k)}\big) \qquad (18)$$

where $\boldsymbol{e}\big(\boldsymbol{\Theta}^{(k)}\big)$ is the vector which represents all the errors, $\boldsymbol{J}^T(\boldsymbol{\Theta}^{(k)})$ is the Jacobian matrix related to function $\boldsymbol{e}(\boldsymbol{\Theta})$ and $\mu_k$ is a value which depends by the eigenvalues of $\boldsymbol{J}^T\big(\boldsymbol{\Theta}^{(k)}\big)\boldsymbol{J}\big(\boldsymbol{\Theta}^{(k)}\big)$. Note that when $\mu_k$ approaches to zero the algorithm becomes Gauss-Newton.

The numerical implementation of the Levenberg-Marquardt algorithm is based on the backpropagation procedure in which derivatives are computed from the last layer to the first.

## 2.6 Procedure to generate the training and validation ANN dataset

In order to build a data-driven model with ANN it is mandatory a training and validation dataset, that consists in input and output values of a process. In the present work, two different approaches have been considered: forward ANN that considers the released mass fluxes at the sources as input and concentrations observed at monitoring points as output; inverse ANN that considers concentrations observed at monitoring points as input and released mass fluxes at the sources as output. The purposes of the approaches are different: the forward one predicts concentrations at monitoring points starting from a known mass flux release at the source, while the inverse one estimates the released mass flux at the source starting from known concentrations observed at the monitoring points. In general, the training and validation dataset, for data-driven models, can be generated by field data or through the results of a numerical model. In this case, the synthetic example of Ayvaz (2010) has been considered and the dataset has been generated through a numerical model built by means of MODFLOW (Harbaugh, 2005) and MT3D (Zheng and Wang, 1999). The procedure applied to generate the dataset consists of the following steps:

1. building of groundwater flow and transport numerical models that reproduce the studied aquifer;

2. definition of the mass released at the source;

3. multiple execution of forward flow and transport models in order to compute the concentrations at the monitoring points.

The crucial point is the definition of the size of the dataset and of the extreme values of the mass released. The size of the training and validation dataset (which corresponds to the number of forward simulations of the numerical model) is defined based on the complexity of the scenario considered. Moreover, the size is usually defined in order to reduce as much as possible the computational cost and, at the same time, in such a way that the dataset itself satisfies the training and validation process of the network. To reduce the number of input dataset, required for the training and validation, Latin Hypercube Sampling (LHS) was considered. LHS randomly generates variables, sufficiently equally distributed and, at the same time, strongly not correlated, from a multidimensional distribution. For forward ANNs the range of the mass released should covers the available mass release data. For inverse ANNs, knowing the concentrations observed at monitoring wells ($C_{true}$), the extremes could be defined through a preliminary run of the numerical flow and transport models that allows to evaluate the relationship between source and observations. The procedure consists in injecting a constant mass rate ($M_0$) at the source, observing the maximum concentrations at monitoring wells ($C_{max}$) and computing the ratio between $C_{true}$ and the maximum concentrations computed by the numerical model ($R = C_{true}/C_{max}$) by injecting $M_0$. Approximating with a linear relationship between the mass release and concentrations at monitoring wells it is possible to define the upper limit of the input dataset as a value greater than $M_0 \cdot R$. The lower limit is 0.

## 2.7 Evaluation of performance

In order to have a comparison on the results obtained by the data-driven model and that obtained by the physical model, the metrics have been defined according to those used in Ayvaz (2010) and Jamshidi et al. (2020): normalized error (*NE*), percent average estimation error (*PAEE*), standard deviation (*SD$_t$*), mean error (*ME*), mean absolute error (*MAE*), root mean squared error (*RMSE*) and normalized root mean squared error (*NRMSE*). The metrics are defined as follow:

$$NE(\%) = \frac{\sum_{i=1}^{M}|\hat{Z}_i - Z_i|}{\sum_{i=1}^{M} Z_i} \cdot 100 \tag{19}$$

$$PAEE_i(\%) = \frac{|\hat{Z}_i - Z_i|}{Z_i} \cdot 100 \tag{20}$$

$$SD_t = \sqrt{\frac{\sum_{r=1}^{N_R}(\hat{Z}_{t,r} - \bar{\hat{Z}}_t)^2}{N_R - 1}} \tag{21}$$

$$ME = \frac{\sum_{i=1}^{M}(\hat{Z}_i - Z_i)}{M} \tag{22}$$

$$MAE = \frac{\sum_{i=1}^{M}|\hat{Z}_i - Z_i|}{M} \tag{23}$$

$$RMSE = \sqrt{\frac{\sum_{i=1}^{M}(\hat{Z}_i - Z_i)^2}{M}} \tag{24}$$

$$NRMSE(\%) = \frac{RMSE}{(Z_{max} - Z_{min})} \cdot 100 \tag{25}$$

where $M$ is the number of unknowns, $Z_i$ is the actual observed value (concentration observed at monitoring points for the forward problem and mass flux released at the source for the inverse one) $\hat{Z}_i$ is the estimated value, $\hat{Z}_{t,r}$ is the estimated value at time $t$ and realization $r$, $\bar{\hat{Z}}_t$ is the estimated value at time $t$ averaged on $N_R$ realizations. $Z_{max}$ and $Z_{min}$ are respectively the maximum and minimum actual observed value.

## 2.8 Study case

To assess the reliability of the proposed approach, a literature case study, presented by Ayvaz (2010) and adopted by Xing et al. (2019) and Jamshidi et al. (2020), has been considered. Firstly, the model has been set up according to Ayvaz (2010). The domain of the solution is a network with block-centered grids. Fig. 2 shows the discretization grid of the numerical model of the domain. Table 1 summarizes the hydraulic and geometry characteristics. Concerning the boundary conditions, the specified head boundary conditions on the upper-left (A-B) and lower-right (C-D) have been defined, otherwise no-flow boundary conditions has been considered for the remainder of the domain. The aquifer is structured in five zones (Fig. 2) with different hydraulic conductivity (HK) values: $HK_1 = 0.0004$ m/s, $HK_2 = 0.0002$ m/s, $HK_3 = 0.0001$ m/s, $HK_4 = 0.0003$ m/s and $HK_5 = 0.0007$ m/s. Hydraulic conductivity is uniform in each zone; therefore, the flow conditions are defined as steady-state and non-uniform. Inside the aquifer, there are two active sources and seven monitoring locations. The simulations cover a total period of five years divided into ten stress periods (six-months each). The sources are active for the first two years of simulation, releasing a conservative contaminant (golden-test) as proposed by Ayvaz (2010). In this regard, the contaminant transport process is transient.

The longitudinal and transverse dispersivity coefficients $\alpha_L$ and $\alpha_T$ are related to the longitudinal and transversal dispersion components $D_L$ and $D_T$ of the dispersion tensor $\boldsymbol{D}$ by the following relationship $D_L = \alpha_L u$ e $D_T = \alpha_T u$, where $u$ is the effective velocity of the flow field.

The data-driven model has been trained and validated for solving two different study cases: forward and inverse transport problems. For the forward approach one scenario has been developed: estimation of the contaminant in terms of concentration values due to the action of two sources at a known position (FWD 1). For the inverse approach different scenarios have been proposed: estimation of the release history of one source with known position (INV 1), estimation of the release history and the location of one source with unknown position (INV 2), estimation of the release histories of two sources with known positions (INV 3) and estimation of the release histories of two sources with known positions together with the order of magnitude of the observations error (INV 4).

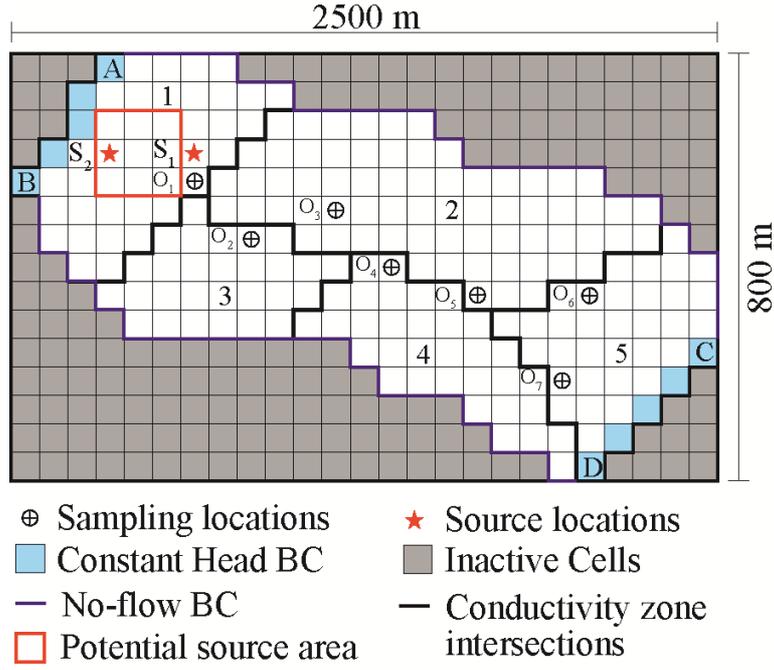

Fig. 2. Discretization grid of the two-dimensional aquifer

Table 1 - Hydraulic and geometry characteristics of the study domain

| Parameters | Values |
|---|---|
| Effective porosity, $\phi$ | 0.3 |
| Longitudinal dispersivity, $\alpha_L$ (m) | 40 |
| Transverse dispersivity, $\alpha_T$ (m) | 4 |
| Saturated thickness, $b$ (m) | 30 |
| Grid spacing in the $\zeta$ direction, $\Delta\zeta$ (m) | 100 |
| Grid spacing in the $\eta$ direction, $\Delta\eta$ (m) | 100 |
| Length of the stress periods, $\Delta t$ (months) | 6 |
| Initial concentration (ppm) | 0 |

## 2.9 Training and validation of the network

Before training the network, as a preprocessing phase of the dataset, the *mapminmax* (MATLAB, 2021) function has been used to rearrange the input $\{x^{(i)}\}_{i=1,...N}$ and the output $\{y^{(i)}\}_{i=1,...N}$ dataset into $\{\tilde{x}^{(i)}\}_{i=1,...N}$ and $\{\tilde{y}^{(i)}\}_{i=1,...N}$ with values in the range [-1 1] according to the following formulas performed on each component $j \in \{1, ..., d_1\}$ and $h \in \{1, ..., d_3\}$:

$$\tilde{x}_j^{(i)} = 2\left(\frac{x_j^{(i)} - \min_{k=1,...,N}\{x_j^{(k)}\}}{\max_{k=1,...,N}\{x_j^{(k)}\} - \min_{k=1,...,N}\{x_j^{(k)}\}}\right) - 1 \qquad (26)$$

$$\tilde{y}_h^{(i)} = 2\left(\frac{y_h^{(i)} - \min_{k=1,\dots,N}\{y_h^{(k)}\}}{\max_{k=1,\dots,N}\{y_h^{(k)}\} - \min_{k=1,\dots,N}\{y_h^{(k)}\}}\right) - 1 \qquad (27)$$

Then, during the training and validation phases of the network, "control" criteria, to evaluate the performance of the two phases, have been set up. Firstly, the maximum number of training epochs has been defined equal to 1000. Secondly, to avoid overfitting, the number of validation checks has been fixed to 6. Table 2 describes the input and output dataset used for all the scenarios investigated. For each scenario analyzed, 70% of the dataset was used to train the network while the remaining 30% for validation.

Table 2 - Summary of the input-output data for the investigated scenarios

| Scenario | Size of the train/validation dataset | Input data | Output data |
|---|---|---|---|
| FWD1 | 500 | Mass release at 2 sources for 2 years every 6 months: 8 data | Concentrations at 7 monitoring points observed for 5 years, one time per year: 35 values. |
| INV1 | 256 | Concentrations at 7 monitoring points observed at time 5 years after the release: 7 data | Mass release at one source for 2 years every 6 months: 4 data |
| INV2 | 2304 | Concentrations at 7 monitoring points observed at time 5 years after the release: 7 data | Mass release at one source for 2 years every 6 months. Planar coordinates of the source: 6 data |
| INV3 | 500 | Concentrations at 7 monitoring points observed for 5 years, one time per year: 35 data, reduced to 26 | Mass release at 2 sources for 2 years every 6 months: 8 data |
| INV4 | 500 | Concentrations at 7 monitoring points observed for 5 years, one time per year: 35 data, reduced to 26 | Mass release at 2 sources for 2 years every 6 months. Error on observations: 9 data |

## 2.10 Forward approach

### 2.10.1 Two sources with known positions (FWD 1)

In this scenario, the goal was to obtain the concentration values at the observation points for each year for a total of 35 values (7 monitoring points for 5 years of simulation). A dataset of 500 input data (six-month release history) and 500 output data (concentrations values at each year) has been used to train and validate the network (Table 2). For the testing phase, reference is made to the concentration values tested in the original Ayvaz (2010) case study. The input layer is the real linear space of dimension eight composed by eight neurons which represent the six-month release history values of the two sources. The hidden layer is composed by 10 neurons activated by the hyperbolic tangent function. The output layer is the real linear space of dimension thirty-five composed by 35 neurons which represent the concentrations values observed in the monitoring locations for each simulation year. The activation function used between the hidden and the output layers is the identity function. For the weights correction procedure, the Levenberg-Marquardt algorithm has been used.

## 2.11 Inverse approach

### 2.11.1 One source with known position (INV 1)

A dataset of 256 input data (concentrations at the time $t = 5$ years) and 256 output data (six-month release history) has been used to train and validate the network (Table 2). The golden-test has been used to assess the good result of the training and validation phases. The input layer is the real linear space of dimension seven composed by seven neurons which represent the seven concentration values at monitoring locations at the time $t = 5$ years. The hidden layer consists of 10 neurons activated by the hyperbolic tangent function. The output layer is the real linear space of dimension four composed by four neurons which represent the four six-month release history values. The activation function used between the hidden and the output layers is the identity function. For the weights correction procedure, the Levenberg-Marquardt algorithm has been used.

### 2.11.2 One source with unknown position (INV 2)

In this scenario, the goal is to identify the release history of the Source 2 together with the location, by estimating the coordinates $\zeta$ and $\eta$. Nine possible cells of the study domain in which the source can be located have been selected. Each cell is identified by its $\zeta$ and $\eta$ coordinates. These values will be returned as output from the network, along with the released mass rate values of the six-month release history. A dataset of 2304 input data (concentrations at the time $t = 5$ years) and 2304 output data (six-month release history and coordinates of the source) has been used to train and validate the network (Table 2). The size of the dataset equal to 2304 derives from using nine dataset of size 256 referred to each hypothetical location of the source within the domain. The type and structure of the network are the same as described for INV1, the only difference is that the output layer is the real linear space of dimension six composed by six neurons which represent the four six-month release history values and the $\zeta$ and $\eta$ coordinates. The same activation functions between the layers and the same learning algorithm have been used.

**2.11.3 Two sources with known positions (INV 3)**

In this application a dataset of 500 input data (concentrations) and 500 output data (six-month release history) has been used to train and validate the network (Table 2). Concerning the testing phase, the six-month release history values tested in the original Ayvaz (2010) case study are used in this work. The input data are no longer the concentration values at the monitoring locations at time $t = 5$ years, but they are the concentration values recorded each year at the observation points, for a total of 35 values (7 monitoring points for 5 years of simulation). Considering that 9 out of 35 monitoring points present, independently of the mass released, values close to zero, the input data was reduced to 26 values for the 500 synthetic simulations. Therefore, the input layer is the real linear space of dimension 26 composed by 26 neurons. The hidden layer consists of 10 neurons activated by the hyperbolic tangent function. Since there are two sources, the output layer is the real linear space of dimension eight composed by eight neurons which represent the four six-month release history values for the two sources within the domain. Again, for the weights correction procedure the Levenberg-Marquardt algorithm has been used.

**2.11.4 Two sources with unknown observation error (INV 4)**

The last application aims to estimate the release history of the two sources with known position together with the estimation of the order of magnitude of the observations error. The training dataset is the same used for INV3, with the exception of the output dataset which included also the value describing the order of magnitude of the observations error (Table 2). Therefore, the output layer is the real linear space of dimension nine composed by eight neurons which represent the four six-month release history values for the two sources within the domain and the order of magnitude of the observation error.

# 3 Results

In this Section, the results related to the forward and inverse approaches of the data-driven models are described. All the results, according to the metrics defined in Section 2.7, were obtained as the average of the outputs produced by 10 neural networks trained with the same dataset. For each run of the network, the dataset is randomly divided into training and validation sets. Furthermore, the initialization of the weights represents another random process within the training and validation procedure. According to this, the 10 neural networks provide modestly different results. Mainly, the choice to work with 10 networks is related with the aim to identify, together with the desired output, the uncertainty of the results. In this regard, in the metrics defined in Section 2.7, the Eq. 21 indicates the confidence interval in which the processed output, obtained as the average of the 10 realizations, falls. Furthermore, if the number of parameters of the network, a priori dependent on the choice of its architecture, is comparable with the total number of points in the training set or if the data is noisy, one way to prevent overfitting is to train multiple neural networks and average their outputs.

## 3.1 Forward

The results produced by the data-driven model concerning the FWD 1, well suit with the one processed by the numerical model. The FWD 1 scenario represents the only scenario in which the data-driven model replaced the numerical model as a surrogate. The estimated concentrations denote a good agreement with the actual values, as shown in Fig. 3. In Table 3, the computed metrics on concentrations at monitoring wells, shows that the difference between estimated and the real values is negligible, highlighting the success of the network. Furthermore, the ANN response is extremely fast (0.2 seconds), while MODFLOW + MT3DMS require about 6 second to run.

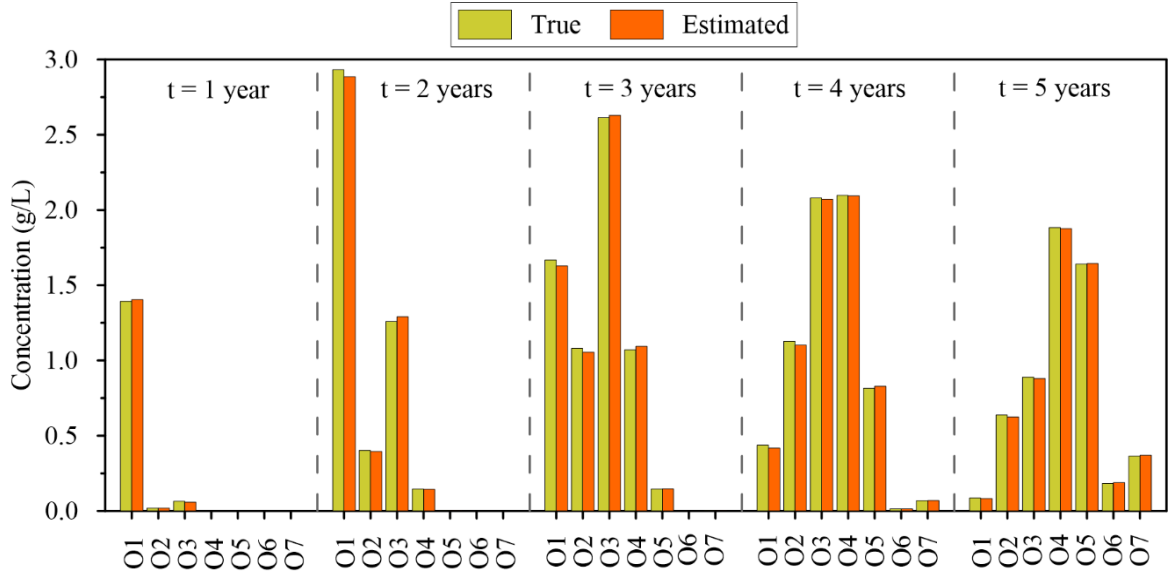

Fig. 3. Observed and estimated concentration at 7 monitoring wells for 5 years of simulation recorded one time per year, forward simulation with two release sources
(FWD 1)

Table 3: ME, MAE, RMSE, and NRMSE computed on concentrations (mg/l) and related to the 35 average concentration values of the 7 monitoring wells (FWD 1)

| | |
|---|---|
| ME (g/L) | 0.0028 |
| MAE (g/L) | 0.0096 |
| RMSE (g/L) | 0.0153 |
| NRMSE | 0.52% |

## 3.2 Inverse

### 3.2.1 One source with position and with concentrations in monitoring points known (INV 1)

The first application INV 1 deals with an inverse simulation with one release source at known position, concentrations in monitoring locations $\xi$ and different error level. The error is processed as:

$$C_{error}(\xi,t) = C_{real}(\xi,t) + \alpha \varepsilon C_{real}(\xi,t) \tag{28}$$

where $\varepsilon$ is a random value obtained from a Gaussian standard distributions and $\alpha$ is the order of amplitude of the error. Normal random errors equal to 0.1%, 1%, and 10% of the standard deviation have been tested in these applications. In Fig. 4 the results obtained for the error-free data and corrupted data with $\alpha = 0.1$ are reported. For both sets of error, the result appears to be quite similar and in accordance with the true release history (golden-test). Also, for this inverse application, the ANN is totally able to well recognize the desired output. As evidence, in Table 4 the observed and estimated source release histories, together with the computed metrics for the different error level are reported, while in Table 5 the metrics for all the error level are described.

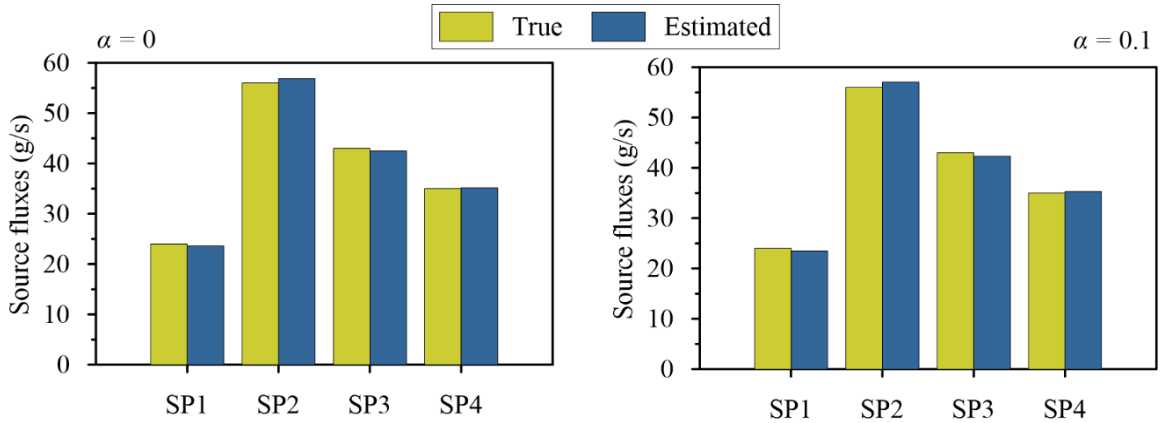

Fig. 4. Observed and estimated release obtained as average of the results of 10 neural networks at known source, inverse simulation with one release source and different error level, error-free data ($\alpha = 0$) and corrupted data ($\alpha = 0.1$) (INV 1)

Table 4: Observed and estimated source release histories (g/s) obtained as average of the results of 10 neural networks with related metrics NE, PAEE, SD for different error level, error-free data ($\alpha = 0$) and corrupted data ($\alpha = 0.1$) (INV 1)

| Source | Stress period | Actual source fluxes (g/s) | $\alpha = 0$ | | | | $\alpha = 0.10$ | | | |
|---|---|---|---|---|---|---|---|---|---|---|
| | | | Average estimated source fluxes (g/s) | NE (%) | PAEE (%) | $SD_t$ (g/s) | Average estimated source fluxes (g/s) | NE (%) | PAEE (%) | $SD_t$ (g/s) |
| $S_2$ | 1 | 24 | 23.61 | 1.22 | 1.65 | 0.39 | 23.48 | 1.63 | 2.18 | 0.32 |
| | 2 | 56 | 56.88 | | 1.58 | 0.75 | 57.07 | | 1.92 | 0.92 |
| | 3 | 43 | 42.52 | | 1.12 | 0.65 | 42.33 | | 1.56 | 0.78 |
| | 4 | 35 | 35.16 | | 0.47 | 0.37 | 35.30 | | 0.86 | 0.37 |

Table 5: ME, MAE, RMSE, and NRMSE computed on source fluxes (g/s) described by four stress period and obtained as average of the results of 10 neural networks for different error level (INV 1)

| | $\alpha = 0$ | $\alpha = 0.001$ | $\alpha = 0.01$ | $\alpha = 0.10$ |
|---|---|---|---|---|
| ME (g/s) | -0.04 | -0.02 | -0.05 | -0.04 |
| MAE (g/s) | 0.48 | 0.34 | 0.53 | 0.64 |
| RMSE (g/s) | 0.55 | 0.42 | 0.63 | 0.70 |
| NRMSE | 1.71% | 1.32% | 1.97% | 2.19% |

### 3.2.2 One source with position unknown and with concentrations in monitoring points known

In inverse application INV 2, although the described problem is more complex than the previous one, the results achieved by the data-driven model are quite satisfactory. In this case, the observations for scenario INV 2 are the same used for the scenario INV 1, but unknows have been risen from 4 (4 releases, INV 1) to 6 (4 releases and source coordinates, INV 2). For this reason, the output reproduced by the ANN is not accurate as the output reproduced for the INV 1 scenario. However, by increasing the number of observations and, consequently, the number of information entering the neural network, the network itself would be able to perform better also in the INV2 scenario. In any case, the ANN manages to estimate quite well not only the release history, but also the location of the source for different set of error level. In Table 6, the actual and estimated coordinates of the source highlight how the network is able to provide well predictions dealing with different errors. Visually, this can be noticed in Fig. 5 where the estimated releases mass rate is compared to the real values, highlighting a good agreement. Table 7 and Table 8 show the computed metrics for the different set of error.

Table 6: Actual and estimated source location ($\zeta, \eta$) obtained as average of the results of 10 neural networks with different data error level, error-free data ($\alpha = 0$) and corrupted data ($\alpha = 0.1$) (INV2)

| Source | Actual location | $\alpha = 0$ | | $\alpha = 0.10$ | |
| --- | --- | --- | --- | --- | --- |
| | | Average estimated location | SD$_t$ (-) | Average estimated location | SD$_t$ (-) |
| $S_2$ | $\zeta = 4$ | $\zeta = 4.02$ | 0.199 | $\zeta = 4.16$ | 0.212 |
| | $\eta = 4$ | $\eta = 3.83$ | 0.178 | $\eta = 3.85$ | 0.271 |

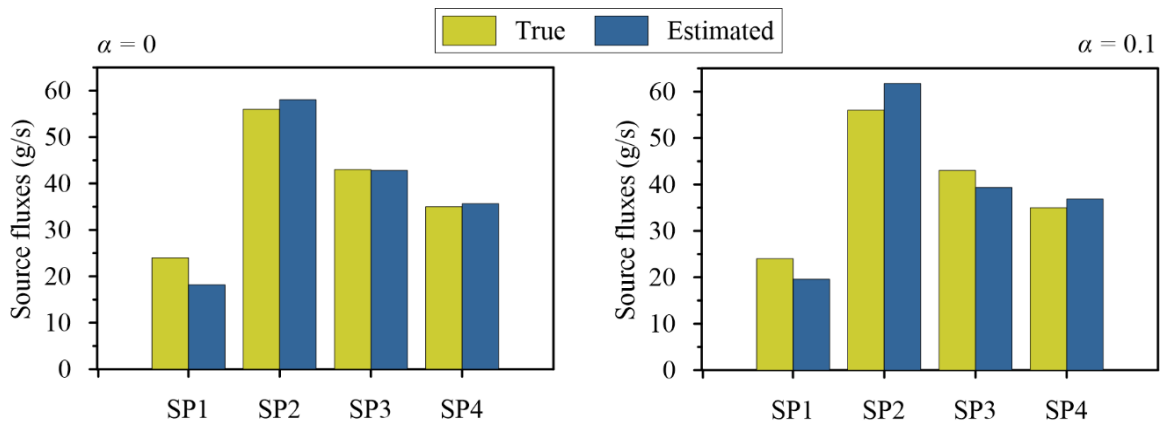

Fig. 5. Observed and estimated release obtained as average of the results of 10 neural networks at estimated unknown source, inverse simulation with one release and different error level, error-free data ($\alpha = 0$) and corrupted data ($\alpha = 0.1$) (INV 2)

Table 7: Observed and estimated source release histories (g/s) obtained as average of the results of 10 neural networks with related metrics NE, PAEE, SD for different error level, error-free data ($\alpha = 0$) and corrupted data ($\alpha = 0.1$) (INV 2)

| Source | Stress period | Actual source fluxes (g/s) | $\alpha = 0$ | | | | $\alpha = 0.10$ | | | |
|---|---|---|---|---|---|---|---|---|---|---|
| | | | Average estimated source fluxes (g/s) | NE (%) | PAEE (%) | $SD_t$ (g/s) | Average estimated source fluxes (g/s) | NE (%) | PAEE (%) | $SD_t$ (g/s) |
| $S_2$ | 1 | 24 | 18.21 | 5.54 | 24.13 | 3.98 | 19.59 | 9.92 | 18.41 | 6.40 |
| | 2 | 56 | 58.09 | | 3.74 | 8.31 | 61.73 | | 10.24 | 10.67 |
| | 3 | 43 | 42.80 | | 0.47 | 7.08 | 39.33 | | 8.54 | 9.99 |
| | 4 | 35 | 35.67 | | 1.91 | 8.53 | 36.85 | | 5.27 | 7.43 |

Table 8: ME, MAE, RMSE, and NRMSE computed on source fluxes (g/s) described by four stress period and obtained as average of the results of 10 neural networks for different error level (INV 2)

| | $\alpha = 0$ | $\alpha = 0.001$ | $\alpha = 0.01$ | $\alpha = 0.1$ |
|---|---|---|---|---|
| ME (g/s) | 0.81 | 1.80 | 0.58 | 0.13 |
| MAE (g/s) | 2.19 | 2.07 | 4.32 | 3.92 |
| RMSE (g/s) | 3.10 | 3.46 | 4.83 | 4.16 |
| NRMSE | 9.69% | 10.81% | 15.11% | 13.01% |

### 3.2.3 Two sources with position and with concentrations in monitoring points known

Scenario INV 3 deals with the estimation of the release history of two sources with known position by means of concentrations in monitoring points of the domain and different level of error. The results obtained by the data-driven model have been compared to those produced by other two literature study cases, example 2 of Ayvaz (2010) and first case of Jamshidi et al. (2020), in order to assess the reliability of the neural network. Although the results previously obtained from the two literature studies are absolutely valid, the neural network is capable to estimate the release histories with greater precision. In fact, by way of example, referring to the metrics shown in Table 9, the NE (%) value obtained in the present work, for the corrupted error level $\alpha = 0.1$, is equal to 1.23% that is much lower if compared with the values obtained in Ayvaz (2010) and Jamshidi et al. (2020) which are respectively 8.06% and 18.06%. In Fig. 6 the results of the estimated release histories, for the error free level $\alpha = 0$ and the corrupted error level $\alpha = 0.1$, are shown, highlighting the agreement between the ANN prediction and the actual values. A final purely visual comparison is shown in Fig.

7 where it is possible to notice that the release histories reproduced through the neural network are better suited to describe the real ones. Furthermore, the error bars related to one time the standard deviation were reported in order to highlight the reliability of the results. Table 10 shows a comparison of statistical metrics between literature cases and the present work for different error level.

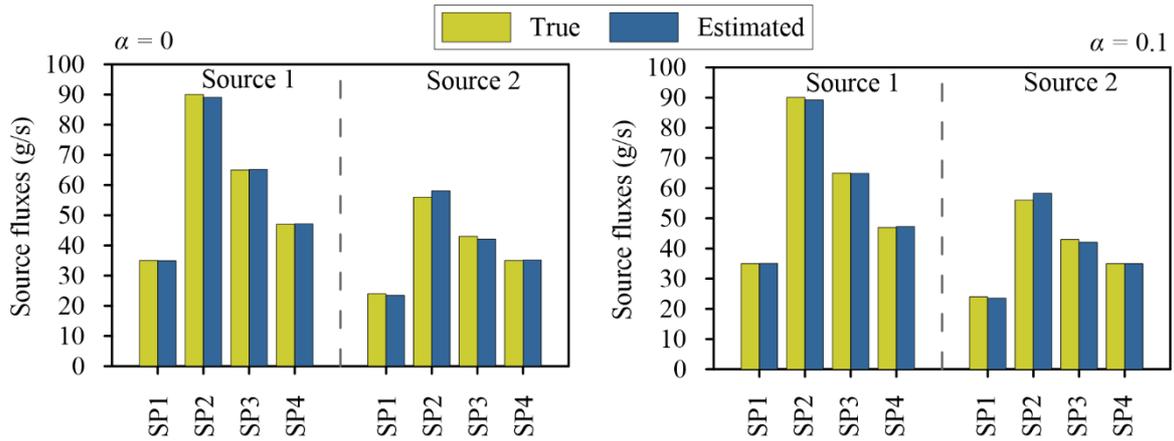

Fig. 6. Observed and estimated release described by four stress period and obtained as average of the results of 10 neural networks results, inverse simulation with two release sources and different error level, error-free data ($\alpha = 0$) and corrupted data ($\alpha = 0.1$) (INV 3)

Table 9: Comparison of the estimated and actual source release histories described by four stress period at two known sources obtained as average of the results of 10 neural networks with Ayvaz (2010), Jamshidi et al. (2020) and the present work, with level error $\alpha = 0.1$ and related statistical metrics (INV 3)

|  | Source |  | $S_1$ | | | | $S_2$ | | | |
|---|---|---|---|---|---|---|---|---|---|---|
|  | Stress period |  | 1 | 2 | 3 | 4 | 1 | 2 | 3 | 4 |
|  | Actual source fluxes | (g/s) | 35 | 90 | 65 | 47 | 24 | 56 | 43 | 35 |
| Ayvaz (2010) | Average estimated source fluxes | (g/s) | 35.4 | 87.5 | 62.9 | 53.4 | 31.5 | 48.5 | 46.9 | 33.6 |
|  | NE | (%) | 8.06 | | | | | | | |
|  | PAEE | (%) | 1.23 | 2.8 | 3.27 | 13.7 | 31.1 | 13.4 | 9.14 | 4.13 |
|  | $SD_t$ | (g/s) | 3.1 | 6.56 | 15.5 | 9.6 | 7.97 | 10.9 | 13.5 | 6.07 |
| Jamshidi et al. (2020) | Average estimated source fluxes | (g/s) | 41.6 | 63.3 | 77.7 | 43.6 | 22.2 | 48.5 | 47.7 | 27 |
|  | NE | (%) | 18.06 | | | | | | | |
|  | PAEE | (%) | 18.9 | 29.6 | 19.5 | 7.15 | 7.6 | 13.4 | 11 | 22.8 |
|  | $SD_t$ | (g/s) | 8 | 29.9 | 42.1 | 23.5 | 11.8 | 35.2 | 42 | 16.9 |
| Present Work | Average estimated source fluxes | (g/s) | 35 | 89.2 | 64.9 | 47.3 | 23.6 | 58.3 | 42.1 | 35 |
|  | NE | (%) | 1.23 | | | | | | | |
|  | PAEE | (%) | 0.05 | 0.9 | 0.15 | 0.69 | 1.76 | 4.09 | 2.06 | 0.01 |
|  | $SD_t$ | (g/s) | 0.17 | 0.43 | 0.34 | 0.34 | 0.29 | 0.81 | 0.79 | 0.27 |

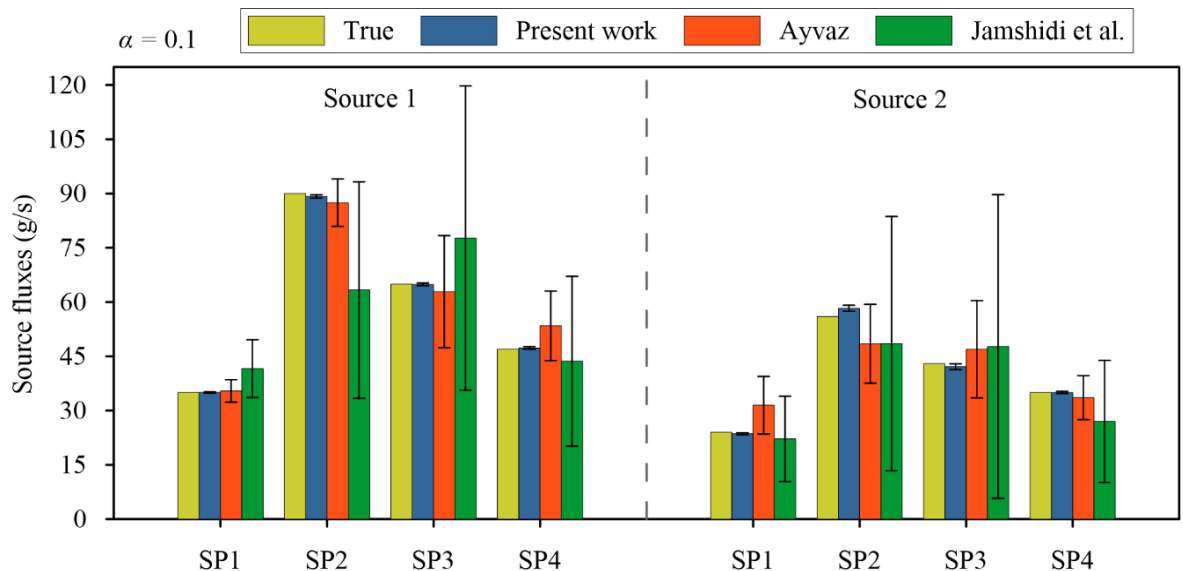

Fig. 7. Estimated release histories in reference works for corrupted data ($\alpha = 0.1$) and, for any time step, the error bars related to one time the standard deviation (INV 3)

Table 10: Comparison of statistical metrics with Ayvaz (2010) and Jamshidi et al. (2020) for different error level, error-free data ($\alpha = 0$) and corrupted data ($\alpha = 0.1$) (INV 3)

|  | $\alpha = 0$ | | | $\alpha = 0.10$ | | |
|---|---|---|---|---|---|---|
|  | **Ayvaz (2010)** | **Jamshidi et al. (2020)** | **Present Work** | **Ayvaz (2010)** | **Jamshidi et al. (2020)** | **Present Work** |
| ME (g/s) | 0.00 | -2.92 | -0.02 | 0.58 | -2.91 | -0.05 |
| MAE (g/s) | 0.85 | 5.65 | 0.63 | 3.98 | 8.92 | 0.61 |
| RMSE (g/s) | 1.06 | 7.34 | 0.90 | 4.77 | 11.58 | 0.93 |
| NRMSE | 1.6% | 11.1% | 1.4% | 7.2% | 17.5% | 1.4% |

### 3.2.4 Simultaneous estimation of the release history and of the error on observations

Scenario INV 4 represents a novelty in the literature: together with the release histories, the order of magnitude of the error on observations has been calculated. The network is able to well recognize the error on observation as shown in Table 11 where the actual and estimated values of the observation errors are reported. Furthermore, as described by Fig. 8 and Table 12, the real and estimated source fluxes for $\alpha = 0$ and $\alpha = 0.1$ are in agreement. In Table 13 the statistical metrics computed on source fluxes, for different estimated error level, are reported.

Table 11: Actual and estimated order of magnitude obtained as average of the results of 10 neural networks of the error on concentrations (INV 4)

| **Actual value** | **Average estimated value** | **SD$_t$ (-)** |
|---|---|---|
| $\alpha = 0$ | $\hat{\alpha} \to 0$ | $\to 0$ |
| $\alpha = 0.001$ | $\hat{\alpha} = 0.00080$ | 0.00011 |
| $\alpha = 0.01$ | $\hat{\alpha} = 0.00996$ | 0.00080 |
| $\alpha = 0.10$ | $\hat{\alpha} = 0.10058$ | 0.00530 |

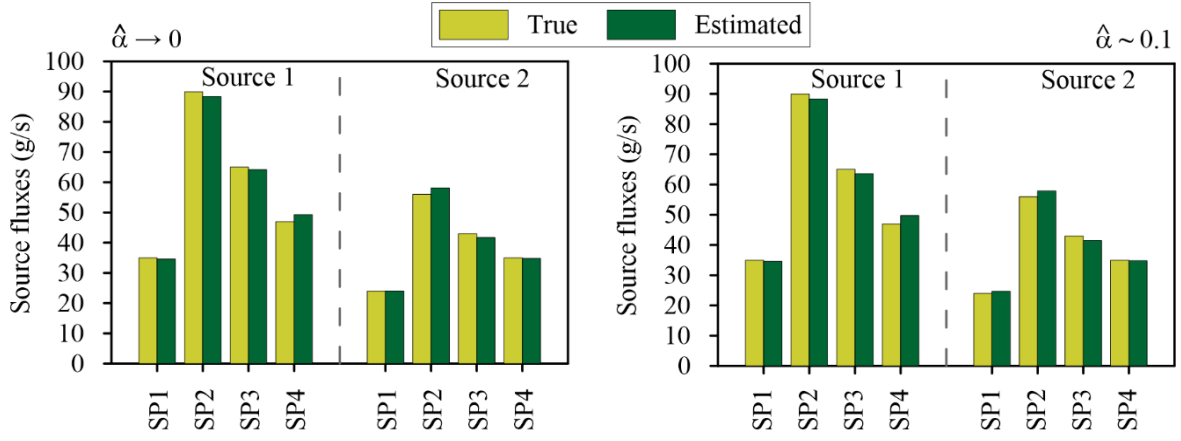

Fig. 8. Observed and estimated release described by four stress period at two known sources obtained as average of the results of 10 neural networks at known source, inverse simulation with two release sources, under different estimated error level, error-free data ($\hat{\alpha} \rightarrow 0$) and perturbated data ($\hat{\alpha} \sim 0.10$) (INV 4)

Table 12: Observed and estimated source release histories (g/s) described by four stress period at two known sources obtained as average of the results of 10 neural networks with related metrics NE, PAEE, SD for different estimated error level, error-free data ($\hat{\alpha} \rightarrow 0$) and perturbated data ($\hat{\alpha} \sim 0.10$) (INV 4)

| Source | Stress period | Actual source fluxes (g/s) | $\hat{\alpha} \rightarrow 0$ | | | | $\hat{\alpha} \sim 0.10$ | | | |
|---|---|---|---|---|---|---|---|---|---|---|
| | | | Average estimated source fluxes (g/s) | NE (%) | PAEE (%) | SD$_t$ (g/s) | Average estimated source fluxes (g/s) | NE (%) | PAEE (%) | SD$_t$ (g/s) |
| $S_1$ | 1 | 35 | 34.59 | | 1.17 | 0.69 | 34.58 | | 1.19 | 0.57 |
| | 2 | 90 | 88.40 | | 1.78 | 0.83 | 88.33 | | 1.86 | 0.76 |
| | 3 | 65 | 64.17 | | 1.28 | 1.42 | 63.63 | | 2.11 | 1.16 |
| | 4 | 47 | 49.29 | 2.24 | 4.86 | 0.67 | 49.69 | 2.65 | 5.72 | 0.54 |
| $S_2$ | 1 | 24 | 24.10 | | 0.38 | 1.15 | 24.71 | | 2.95 | 1.32 |
| | 2 | 56 | 58.14 | | 3.84 | 1.93 | 57.88 | | 3.37 | 2.02 |
| | 3 | 43 | 41.71 | | 2.99 | 1.19 | 41.50 | | 3.48 | 1.26 |
| | 4 | 35 | 34.80 | | 0.58 | 0.67 | 34.79 | | 0.59 | 0.81 |

Table 13: ME, MAE, RMSE, and NRMSE computed on source fluxes (g/s) described by four stress period at two known sources obtained as average of the results of 10 neural networks for different estimated error level (INV 4)

| | $\hat{\alpha} \rightarrow 0$ | $\hat{\alpha} \sim 0.001$ | $\hat{\alpha} \sim 0.01$ | $\hat{\alpha} \sim 0.10$ |
|---|---|---|---|---|
| ME (g/s) | -0.02 | -0.03 | 0.00 | -0.01 |
| MAE (g/s) | 1.11 | 1.11 | 1.27 | 1.31 |
| RMSE (g/s) | 1.37 | 1.36 | 1.57 | 1.52 |
| NRMSE | 2.1% | 2.1% | 2.4% | 2.3% |

# 4 Discussion and Conclusions

In the present work, neural networks have been used as data-driven model to solve different applications related to forward and inverse transport problems, using the concentration values in different monitoring points and release histories as data necessary for the training phase. To summarize, the obtained results show that this data-driven technique is well suited with the aim to provide solutions with very low computational costs for a transport problem that can be useful to the aquifer manger in order to define rapidly remediation strategies. The use of the LHS represents a first advantage of the work carried out, as it allows to reduce the number of forward simulations necessary for the network training, reducing the computational burden. In this regard, Ayvaz (2010) approach minimized an objective function on the basis of the output reproduced by the forward model. This led to a large number of forward simulations to achieve convergence (32,859 simulations). The approach used by Jamshidi et al. (2020) is based on the transfer function theory, with the advantage to run the simulation model only once. The resulting transfer matrices has achieved convergence by means of an optimization algorithm in less than 600 iterations, a comparable number with the simulations processed in the present study. Hence, although the approach described by Jamshidi et al. (2020) is very performing from a computational point of view, the results achieved in the present work are far better. Furthermore, both Ayvaz (2010) and Jamshidi (2020), during the optimization process, use a dataset of observations equal to 140 concentration values. This results in a much lower number of unknowns parameters than the number of measurements. The results obtained through the proposed procedure were carried out by using only 26 observations, substantially limiting the number of a priori information provided to the model. According with the computed metrics for the performance evaluation (Tables 10 and 13), in general, the ANN leads to good results and better if compared with the other two literature cases. For instance, the $SD_t$ values computed for the INV 3 application show a less extensive confidence interval, highlighting that the results obtained as average of the 10 realizations (run of the network) is characterized by a low uncertainty if compared with the $SD_t$ values of the two other studies (Fig. 7). Also, the $SD_t$ values calculated for the INV 4 application highlight the capability of the ANN to deal with the observations error estimation, allowing to manage with a new topic never discussed before in the scientific ANN literature. Established that the results are satisfactorily achieved, the greatest advantage obtained from a data-driven model developed as ANN is the possibility of producing solutions computationally efficient. In fact, once trained, the network will provide the relative outputs without the need to work with the complex numerical model. Neural networks use the forward numerical model only to derive the training dataset, which is defined within a certain range. Therefore, the network, if well trained, will be able to return the desired output. As reported in Chen et al. (2021), one of the most criticisms in literature, is the difficulty of applying these procedures in practice. Usually, the parameters of the aquifer are assumed to be known, when actually they are sparsely known and highly heterogeneous. However, based on the results achieved in the present work, future works will focus on the simultaneous estimation of the network parameters together with the identification of the release history. Furthermore, a topic widely discussed in the scientific community concerns the ability of ANNs to generalize. Neural networks have been shown to be able to generalize, but only within the training range and not outside of it. In this regard, to implement a neural network, there is the necessity to have a priori information in order to define the training intervals suited to the problem to be described and the potential source locations. With the aim to deal with the generalization issue, possible future works may concern the application of a new field, always in the artificial intelligence environment, known as "Physically Informed Neural Networks" (PINNs - Raissi et al., 2019). Without going into detail, this model adds to the usual neural network, information about the physical phenomenon described by general nonlinear partial differential equations (PDEs). In this deep learning method, the physical law becomes a constraint of the network itself as an addition to the objective function to be minimized. The network will be trained and at the same time constrained to the physic that describes the problem. This modeling technique already has achieved good results in the literature (Pang et al., 2019; Mao et al., 2020; Wang et al., 2021) and this motivates the possibility of using it for transport forward and inverse problems which are described by partial differential equations.

# Acknowledgements

This research is supported by the PRIMA programme under grant agreement No1923, project InTheMED. The PRIMA programme is supported by the European Union.